\documentclass{bmvc2k}

\usepackage{amsmath}
\title{Adversarial View-Consistent Learning for Monocular Depth Estimation}

\addauthor{Yixuan Liu\footnotemark[1]}{liuyixua17@mails.tsinghua.edu.cn }{1}
\addauthor{Yuwang Wang}{yuwwan@microsoft.com}{2}
\addauthor{Shengjin Wang}{wgsgj@tsinghua.edu.cn}{1}

\addinstitution{
Tsinghua University\\
Beijing, China 
}

\addinstitution{
 Microsoft Research Asia \\
 Beijing, China
}

\runninghead{Y.LIU, Y.WANG, S.WANG}{AVCL}

\def\eg{\emph{e.g}\bmvaOneDot}

\def\etal{\emph{et al}\bmvaOneDot}
\def\ie{\emph{i.e}\bmvaOneDot}
\def\wrt{\emph{w.r.t}\bmvaOneDot}
\usepackage{amsfonts}
\usepackage{booktabs}
\usepackage{multirow}
\usepackage{amssymb}
\usepackage{color,xcolor}

\begin{document}

\maketitle
\footnotetext[1]{This work was done when Y.Liu was an intern at Microsoft Research Asia}
\begin{abstract}

This paper addresses the problem of Monocular Depth Estimation (MDE). Existing approaches on MDE usually model it as a pixel-level regression problem, ignoring the underlying geometry property. We empirically find this may result in sub-optimal solution: while the predicted depth map presents small loss value in one specific view, it may exhibit large loss if viewed in different directions. In this paper, inspired by multi-view stereo (MVS), we propose an Adversarial View-Consistent Learning (AVCL) framework to force the estimated depth map to be all reasonable viewed from multiple views. To this end, we first design a differentiable depth map warping operation, which is end-to-end trainable, and then propose a pose generator to generate novel views for a given image in an adversarial manner. Collaborating with the differentiable depth map warping operation, the pose generator encourages the depth estimation network to learn from  hard views, hence produce view-consistent depth maps . We evaluate our method on NYU Depth V2 dataset and the experimental results show promising performance gain upon state-of-the-art MDE approaches.


\end{abstract}

 \begin{figure}
    \centering
    \includegraphics[width=\linewidth]{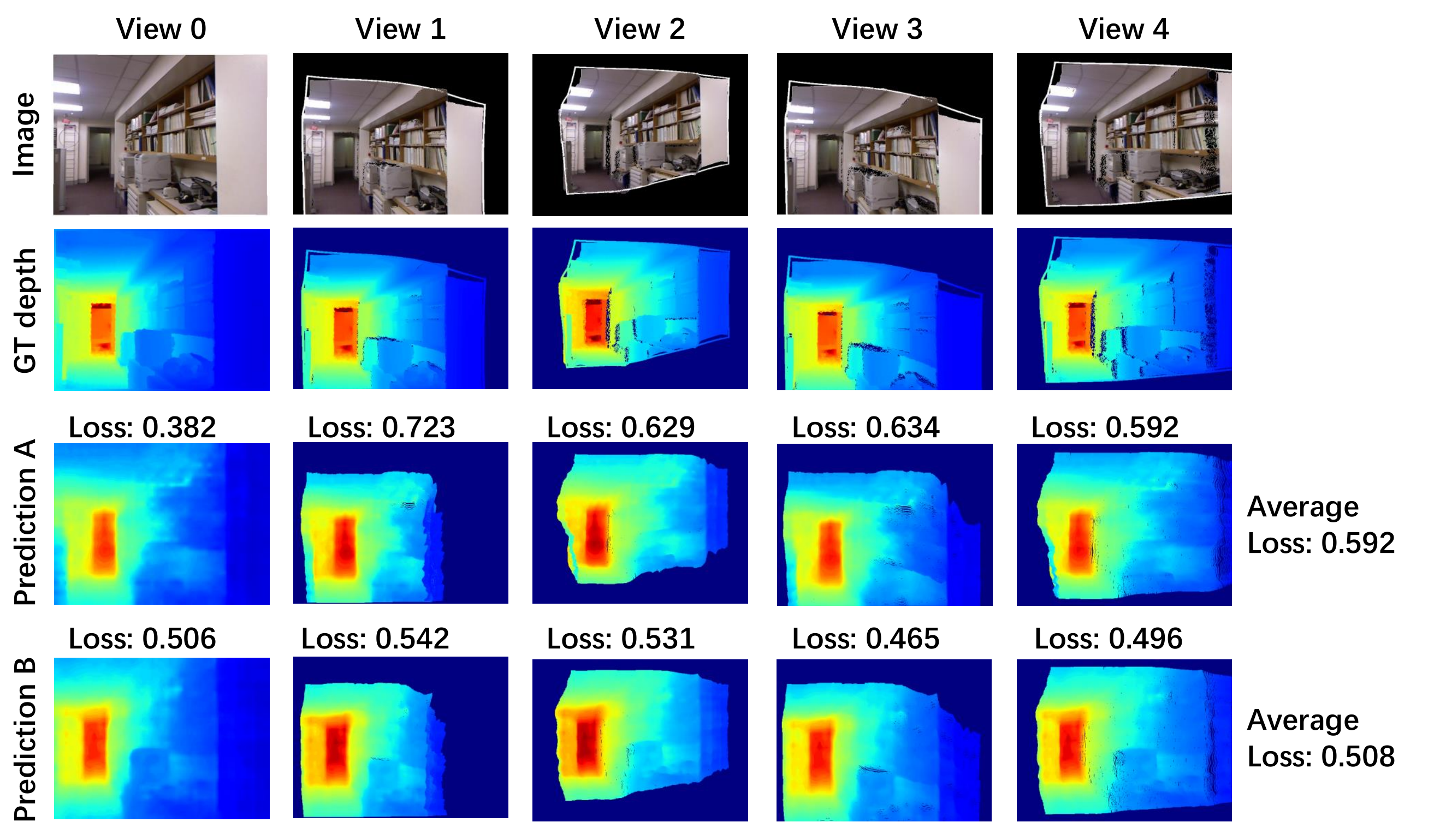}
    \caption{An illustration on view-consistent depth map (prediction B) and view-inconsistent depth map (prediction A). Viewed in multiple direction, prediction B presents small loss value in average. In contract,  prediction A may present small loss  view in some specific view (view 0), but the average loss across multiple views is large.}
    \label{fig:intro}
\end{figure}
\section{Introduction}
\label{sec:intro}

Learning to predict depth from 2D images has been an actively studied problem over recent years, with widespread applications in 3D vision, such as scene modeling, 3D object detection and segmentation. 
Conventional approaches usually require multiple views of the observation or limited environment, so that the application is restricted. 
Therefore, learning to predict depth from a single image, \ie, \emph{Monocular Depth Estimation (MDE)}, is now attracting more research attention.

MDE is a typical ill-posed problem, because one 2D images may correspond to many reasonable 3D scenes. However,  structural prior knowledge of natural images (texture / object shapes / object continuity) can provide rich information to overcome this inherent ambiguity, and is leveraged in early works on MDE~\cite{Make3D, prior1,Ladicky_2014_CVPR,prior2}. 

Recently, along with the rise of deep learning,  MDE performance is significantly facilitated by Deep CNN-based models~\cite{deep1,Wang_2015_CVPR,deep2,crop1,deeperdepth,dorn}, in which the structural prior knowledge is learned from training data. 
Most of the existing CNN-based methods focus on designing more suitable network structures for learning depth~\cite{NIPS2014_5539,deep3d,dorn}, or proposing new learning objectives~\cite{deeperdepth,dorn} to replace the vanilla loss functions (\eg$L_1$ or $L_2$). 
However, no matter with which form of network structure or loss function, depth prediction in every single pixel has equal contribution to the loss value. As a consequence, two depth map prediction with the same loss value, may present quite different results (Fig. \ref{fig:intro}). 

To address this problem. we empirically find that applying multiple supervisions on multiple generated views of the same scene helps the model to learn a more reasonable depth map. Fig.~\ref{fig:intro} shows an illustration: A good depth  prediction, no matter viewed in which direction, always has small loss values with regard to the ground truth, which means the prediction is \emph{view-consistent}. 
In contrast, if a bad depth prediction happens to have small loss in one specific view, it is unlikely the loss values in all views are small as well. In other words, the prediction is not view-consistent.

Moreover, the transformation pose of the view point  also matters during training. With limited training time, it is impractical to enumerate all possible transformation poses. Also, not all views are valuable for training the depth model: if a generated view still has small loss, it means the cross-view regularization does not work much in this view. Therefore, it is of great importance to efficiently  mine \emph{hard} views, which benefit the training more. 

\begin{figure}
    \centering
    \includegraphics[width=\linewidth]{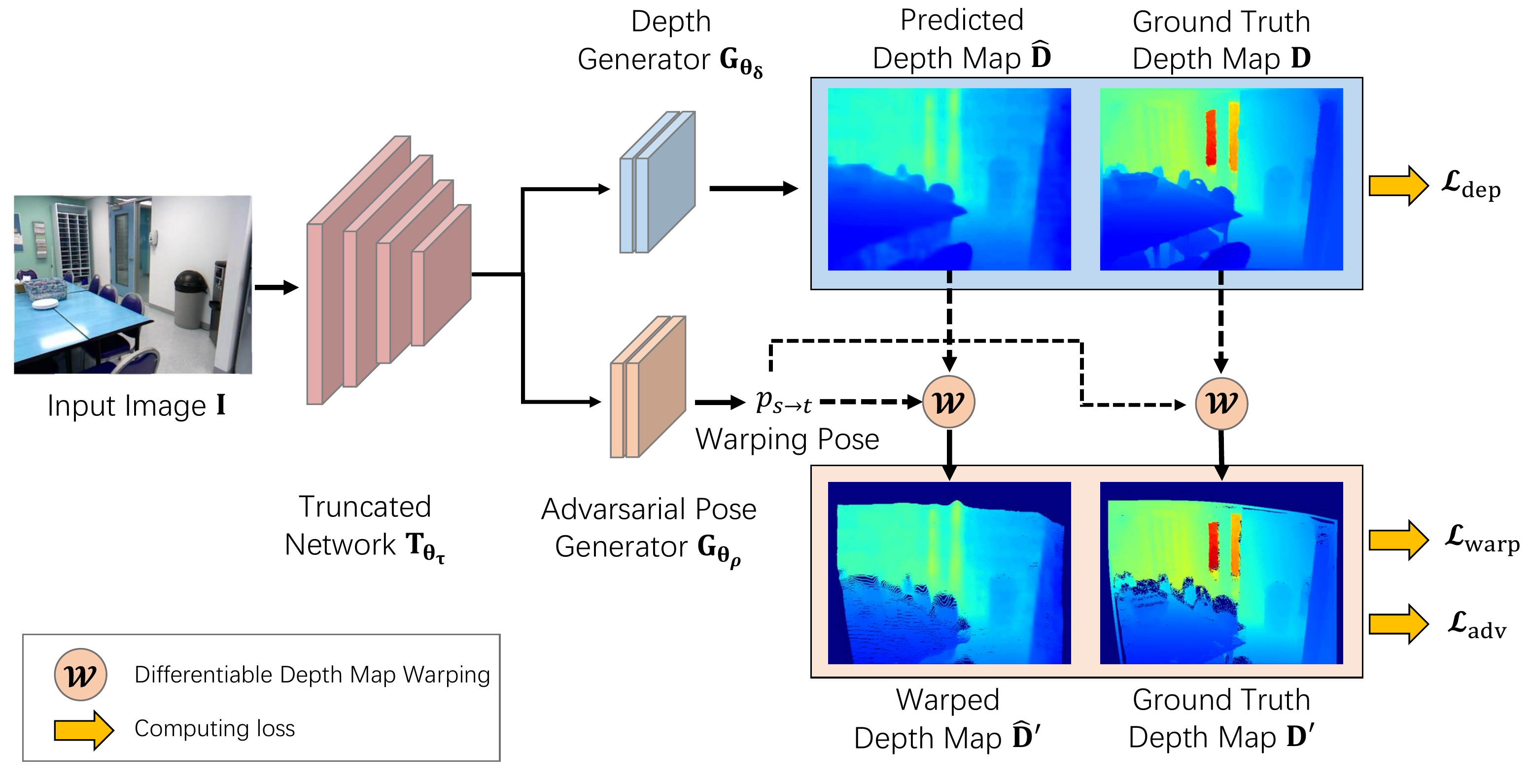}
    \caption{An overview of the proposed Adversarial View-Consistent Learning framework.}
    \label{fig:framework}
\end{figure}

In this work, we propose an Adversarial View-Consistent Learning (AVCL) framework for monocular depth estimation (Fig. \ref{fig:framework}). In AVCL, the predicted depth map for an input image is transformed to another view by a differentiable depth map warping operation, and an additional supervision is applied upon the warped depth map to regularize the predicted depth map to be view-consistent. Furthermore, in order to efficiently mine hard views for the additional supervision, we design a warping pose generation module to automatically generate the warping pose, which is trained adversarially against the objective of the warped branch. Experiments on NYU Depth v2 dataset show both the view-consistent (multiple supervisions) learning and also the adversarial pose generation module bring significant performance gain on baselines in terms of 7 commonly used metrics. Moreover, as AVCL is  complementary to existing methods, the performance gain is addable.

The contribution of this work can be summarized as follows:

(1) We show that applying multiple supervisions on top of multiple views of the depth map leads the model to a better optimum, where the predicted depth maps are view-consistent.

(2) We propose a differentiable depth map warping operation to generate multiple views of the predicted depth map, so that additional supervision can be simply applied on the warped branch, and the entire framework is end-to-end trainable.

(3) We  propose an adversarial warping pose generation module to automatically mine the hard views in view-consistent learning, which further benefits the  learning of depth.


\section{Related Work}
\label{sec:relatedwork}
\textbf{Monocular Depth Estimation}. Early works~\cite{Make3D,singleimage} in MDE usually base on hand-crafted features~\cite{prior2,Ladicky_2014_CVPR,sift} and probabilistic graph models~\cite{Make3D, resolution,discrete-continuous}, learn from  structral prior knowledge of natural images such as texture and object shapes, while suffer in  unconstrained scenes due to excessively strong assumptions on scene geometry.
Recent advances in MDE ~\cite{depthtransfer,whatuncertainty,deeperdepth,Li_2017_ICCV,Liu,deep2,dorn} mostly employ deep CNN to learn depth from large amount of RGB-D data, mitigating previous strong geometry assumptions, and dramatically increase performance. Most of the existing deep MDE approaches focus on either designing adaptive network architectures~\cite{NIPS2014_5539,deep3d}, or re-formulate the learning objective by introducing new loss fuctions~\cite{deeperdepth,dorn}. Although these methods present considerable accuracy, to our knowledge, there are few existing efforts on learning view-consistent depth maps. 

\noindent \textbf{Multi-view stereo (MVS)} methods reconstruct 3D model of the scene from multi-view images~\cite{furukawa2015multi}. Recent methods~\cite{galliani2015massively, tola2012efficient} focus on depth map fusion and achieve competitive results in practice \wrt  early volumetric-based methods~\cite{seitz1999photorealistic, kutulakos2000theory}  . Deep learning methods~\cite{ji2017surfacenet, yao2018mvsnet} shows great success on this task with the large datasets are avaible. Even though single depth estimation task does not belong to MVS, but MVS methods take depth map as an important intermediate to fusion the 3D representation of the target scene. Our method is inspired by MVS since multiviews act as an effective constraints on the geometry of scenes. The error of depth map estimated from one view can be magnified when observed from other viewpoints.


\noindent \textbf{Adversarial learning} is first proposed for generative models~\cite{gan} and has been extensively studied and leveraged in applications like image synthesis~\cite{dcgan,biggan} and manipulation~\cite{disentangled, rotation,mani}. Despite its success on image generation, the idea of adversarial learning is also applied in discriminative models for representation learning, where the mainly usage it to mine or to generate hard training samples. For instance, Wang \etal~\cite{afastrcnn} employ adversarial learning to generate occlusion and spatial transformation on images for object detection; BIER~\cite{bier} adopt adversarial loss to ensure the diversity of embeddings so that the ensemble yields better accuracy. In this work, we follow these ideas and apply adversarial learning to generate hard views for training.

\section{Approach}
\label{sec:approach}
In this section we introduce the framework of the proposed adversarial view-consistent learning (AVCL).
As shown in Fig. \ref{fig:framework}, AVCL involves two key components, namely \emph{differentiable depth map warping} (Sec. \ref{sec:ddmw}) and \emph{adversarial pose generation} (Sec. \ref{sec:apg}).
In differentiable depth map warping, given an initial depth map and a specific warping pose, a novel view of the depth map is generated, and all the operations are differentiable so that back-propagation can be applied. 
Resort to the differentiable warping operation, we are able to impose multiple supervision signals on the generated novel views of the depth map. 
With supervisions on multiple views, the model is expected to output more view-consistent depth map.
However, how to choose the warping pose here remains a problem. 
To this end, we design an adversarial pose generation module, which is also differentiable, and output an adversarial warping pose. Warped with the adversarial pose, the novel view of depth map is expected to have large loss compared to the warped ground truth depth map. 
In other words, the module adversarially mines a \emph{hard} view to train the depth generator.

The combination of a depth generator, the differentiable depth map warping operation and an adversarial pose generation module forms the AVCL framework and will now be described in more detail in the following sections.

\vspace{-0.3cm}
\subsection{Differentiable Depth Map Warping}
\label{sec:ddmw}  
The differentiable depth map warping operation $\mathcal{W}(\mathbf{D_s}, p_{s\rightarrow t})$ takes a source depth map $\mathbf{D_s} \in \mathbb{R}^{H\times W}$ and a 6-DoF pose vector $p_{s\rightarrow t}$ as input and outputs a target depth map $\mathbf{D_t} \in \mathbb{R}^{H\times W}$. 
The target depth map is a novel view with the camera moves with pose $p_{s\rightarrow t}$.

Let $p_s=(x_s, y_s, 1)^T$ denote the homogeneous coordinates of a pixel in the source depth map $\mathbf{D_s}$. For every $p_s$, we calculate the corresponding target coordinate $p_t = (x_t, y_t, 1)^T$ and depth $z_t$ by,
\begin{equation}
    \begin{pmatrix} x_t z_t\\ y_t z_t\\ z_t \end{pmatrix} = K \left (R_{s\rightarrow t} \mathbf{D_s}(p_s) K^{-1} p_s + T_{s\rightarrow t}\right)
\end{equation}
where $\mathbf{D_s}(p_s)$ is the depth value of the pixel, and $K$ denotes the camera intrinsics matrix. $R_{s\rightarrow t} \in \mathbb{R}^{3\times 3}$ and $T_{s\rightarrow t}  \in \mathbb{R}^{3}$ are the rotation matrix and translation vector from the source view to the target view, respectively, calculated from  $p_{s\rightarrow t}$. 
The whole process can be broken down into three procedures.
First, the source pixel coordinates are mapped into world coordinates, \ie, $p_w = \mathbf{D_s}(p_s) K^{-1} p_s$. 
Second, the camera moves with pose $p_{s\rightarrow t}$, so that the world coordinates are transformed to $p_w^{\prime} = R_{s\rightarrow t} p_w + T_{s\rightarrow t}$. 
Finally, the world coordinates are being transformed into a different camera coordinate frame.

Note that the coordinates $p_t = (x_t, y_t, 1)^T$ here are continuous  coordinates, in order to obtain the pixel-wise depth map $\mathbf{D_t}$, we need to fill the depth values into corresponding pixels. 
Different from the cases in STN~\cite{stn} and SfMLearner~\cite{sfmlearner} where bilinear sampling is employed to deal with the one-to-many mapping, 
in our case this is a many-to-one mapping since different coordinates may fall into the same pixel. To ensure the differentiability, we use the following approximation,
\begin{equation}
\begin{split}
\mathbf{D_t}(i,j) = \min_{x, y} \ z(x, y), \quad \forall (x, y) \in \left \{(x_t, y_t) \mid \lfloor x_t \rfloor = i, \lfloor y_t \rfloor = j, (x_t, y_t)\in \mathcal{P}_t \right \}
\end{split}
\end{equation}
where $i$ and $j$ are the discrete pixel coordinates, $\mathcal{P}_t$ is the collection of all the projected target coordinates, and $z(x, y)$ represents the depth value of the point $(x, y)$. We first map the continuous coordinates into discrete ones with the floor operation. 
Accordingly, if multiple points fall into the same pixel, we compare their depth value, and select the minimum value among them as the depth of this pixel. 
This is reasonable because if we see two objects coincide, the front object always occludes the back one. 
Moreover, if a pixel received no depth value, this pixel is marked as ignored and do not participate in computing the loss.

As a summary, the warping operation $\mathcal{W}(\mathbf{D_s}, p_{s\rightarrow t})$ generates a transformed view of the depth map. The computation therein is either matrix manipulation (the projection procedure) or simple value assignment (the coordinate discretization procedure), so that the entire operation is differentiable and able to be embedded in the network for end-to-end training.

\subsection{Adversarial Pose Generation}
\label{sec:apg}

\begin{figure}
    \centering
    \includegraphics[width=0.8\linewidth]{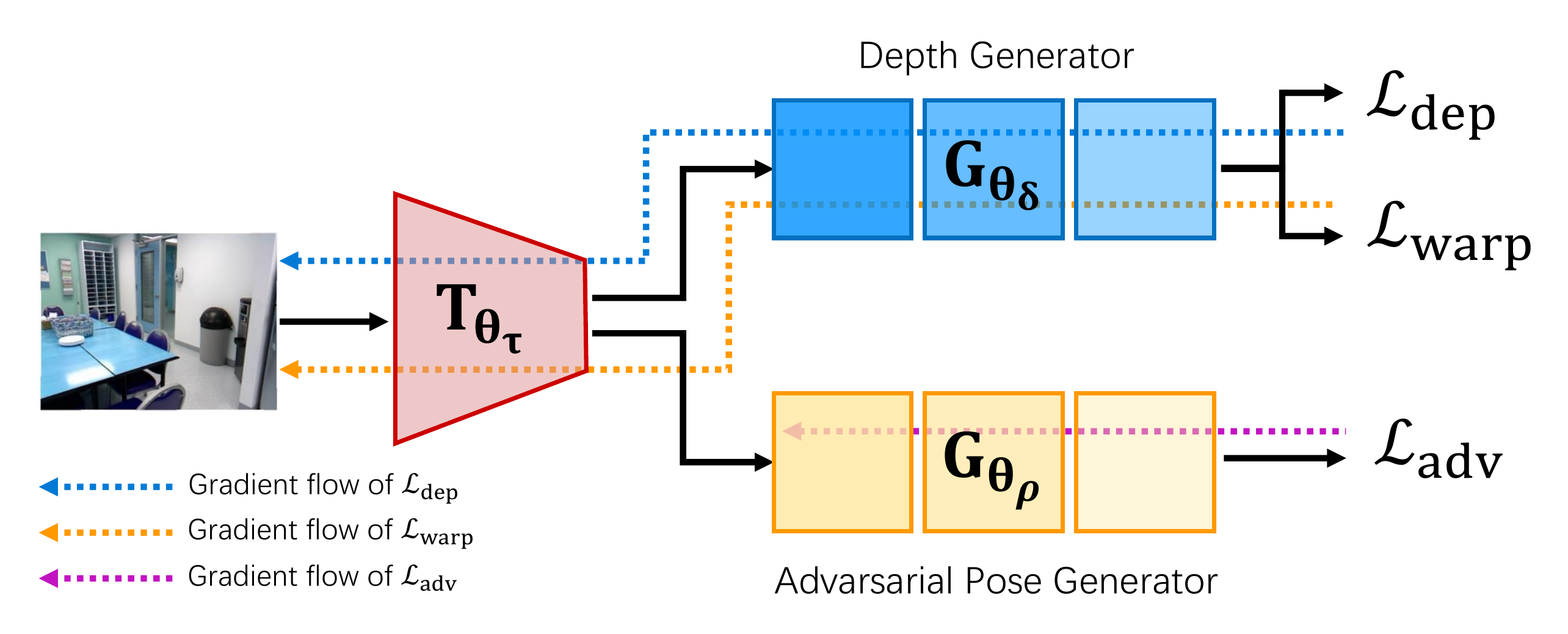}
    \caption{Gradient flow during training in AVCL.}
    \label{fig:grad}
\end{figure}

With the help of the differentiable warping operation, we are able to generate multiple views of the predicted / ground-truth depth map. Accordingly, multiple supervisions can be applied to restrict the depth generator to output more view-consistent depth maps. 
Nevertheless, a question arises here: How should we set the warping pose, so that the model benefits the most from learning such warped depth map? The simplest solution is to randomly sample the warping pose. As will be illustrated in Sec. \ref{exp:results}, random warping pose does help, but the performance gain is limited. 
To address this issue, we propose an adversarial pose generation module to automatically generate the warping pose. The pose generation module is trained adversarially against the depth generator, hence is expected to generate a pose that results in the \emph{hardest} view of the predict depth map. 
In other words, viewed in the generated pose, the predicted depth map is expected to have the largest error \wrt the warped ground truth.

Formally, let us denote the truncated network as $\mathbf{T}_{\theta_{\tau}}(\mathbf{I})$, where $\mathbf{I}$ is the input image, and the output $\mathbf{z} = \mathbf{T}_{\theta_{\tau}}(\mathbf{I})$ is the intermedia feature map. 
The depth generator $\mathbf{G}_{\theta_{\delta}}(\mathbf{z})$, consisting of a $3\times3$ ,an $1\times 1$ convolution layer and a fc layer, takes $\mathbf{z}$ as input and outputs the predicted depth $\mathbf{\hat{D}}$. 
The adversarial pose generation module $\mathbf{G}_{\theta_{\rho}}(\mathbf{z})$ is another branch comprised of a stack of several convolution layers and a fully-connected layer, then followed by a  pre-defined scaled and shifted sigmoid function to restrict the range of the output pose vector $p=\mathbf{G}_{\theta_{\rho}}(\mathbf{z}) \in \mathbb{R}^6$. 
The pose vector $p$ is then used to warp the predicted depth map $\mathbf{\hat{D}}$ and the ground truth depth map $\mathbf{{D}}$, yielding $\mathbf{\hat{D}^{\prime}}$ and $\mathbf{{D}^{\prime}}$ respectively. Accordingly, we compute $L_1$ loss for both $\mathbf{\hat{D}}$ and $\mathbf{\hat{D}^{\prime}}$ ,

\begin{equation}
    \mathcal{L}_{\texttt{dep}} = \Vert \mathbf{\hat{D}} - \mathbf{D} \Vert_1, \quad \mathcal{L}_{\texttt{warp}} = \Vert \mathbf{\hat{D}^{\prime}} - \mathbf{D}^{\prime} \Vert_1
\end{equation}

During back-propagation, we update the parameters in the shared network $\mathbf{T}_{\theta_{\tau}}$ with gradient from both the two branches, $\frac{\partial \mathcal{L}_{\texttt{dep}} + \mathcal{L}_{\texttt{warp}}}{\partial \theta_\tau}$ and update the parameters of the depth generator $\mathbf{G}_{\theta_{\delta}}$ with only $\frac{\partial \mathcal{L}_{\texttt{dep}} }{\partial \theta_\delta}$. 
In contrast, we expect the pose generator to output hard poses, so we optimize the parameters of $\mathbf{G}_{\theta_{\rho}}$ to the opposite direction by inverting the gradient, \ie, using the negative gradient $-\frac{\partial \mathcal{L}_{\texttt{warp}} }{\partial \theta_\rho}$. 
The adversarial loss is only applied to the pose generator and can be written as,

\begin{equation}
    \mathcal{L}_{\texttt{adv}} = - \mathcal{L}_{\texttt{warp}} + \mathbb{\lambda} \cdot (p\odot p)
\end{equation}

where $\mathbb{\lambda}$ is a vector of hyperparameters that regularize the value of the warping pose not to be too large. For an intuitive understanding, we also visualize the gradient flows during back-propagation, and the visualization is shown in Fig \ref{fig:grad}.

Note that the adversarial pose generation module is only used in the training phase. During inference, the pose generation module and warping procedure are discarded and only a single-pass depth generator is used.


\section{Experiments}
\label{sec:exp}
\subsection{Experiment Setup}
\textbf{Dataset.} We adopt NYU Depth v2 dataset~\cite{nyuv2} for both training and evaluation. 
NYU v2 dataset is a large-scale indoor scene understanding dataset that contains about 120K images and corresponding depth maps captured by RGB-D cameras. The testing data consists of 694 images from 215 different scenes. For training, we explore two settings, namely \texttt{all}, which means all the training data (120K) are used, and \texttt{down10}, which means we down-sample the training set by 10 times so that the total number of training data is  $1.2$K.

\begin{table}[]
\footnotesize
    \centering
    \begin{tabular}{c|c|c|ccc|cccc}
    \toprule
         \multirow{2}{*}{Method}&\multirow{2}{*}{\hspace{-0.2cm} AVCL \hspace{-0.2cm}} & \multirow{2}{*}{\hspace{-0.2cm}Data\hspace{-0.2cm}} &\multicolumn{3}{c|}{Higher is better}  & \multicolumn{4}{c}{Lower is better}  \\
    \cline{4-10}
         &&& \scriptsize{$\delta < 1.25$}  \hspace{-0.2cm}& \hspace{-0.2cm}\scriptsize{$\delta < 1.25^2$}\hspace{-0.2cm} &\hspace{-0.2cm} \scriptsize{$\delta < 1.25^3$} \hspace{-0.2cm}&
       \scriptsize{rel}\hspace{-0.2cm}& \hspace{-0.2cm}\scriptsize{log$_{10}$}\hspace{-0.2cm} & \hspace{-0.2cm}\scriptsize{rms}\hspace{-0.2cm}& \hspace{-0.2cm}\scriptsize{rms$_{10}$}\hspace{-0.2cm} \\
    \hline
    L1 Loss~ & &\texttt{down10}
    &0.800	& 0.951& 0.987	& 0.147& 0.061& 0.512& 0.185\\
    berHu Loss~\cite{deeperdepth}  && \texttt{down10}
    & 0.811	& 0.955& 0.988	& 0.143&0.060 & 0.504 & 0.180\\
    DORN Loss~\cite{dorn}  && \texttt{down10}
    &0.817&	0.959&	0.989&	0.138& 0.059 &	0.523&	0.179\\
    \hline
    L1 Loss~ & \checkmark &\texttt{down10}
    &0.810	&0.959  &0.989	&0.141 & 0.060&0.525&0.182 \\
    berHu Loss~\cite{deeperdepth}  & \checkmark & \texttt{down10}
    & 0.822	& 0.960 & 0.989	& 0.121&0.059 & 0.515& 0.177\\
    DORN Loss~\cite{dorn}  & \checkmark & \texttt{down10}
    & 0.827&	0.963&	0.990&	0.120&0.053	&	0.505&	0.176\\
    \hline
    DORN Loss~\cite{dorn}  &  & \texttt{all}
    &0.828&	0.965&	\textbf{0.992}&	0.115&	\textbf{0.051}&	0.509&	0.176\\
    DORN Loss~\cite{dorn}  & \checkmark & \texttt{all}
    &\textbf{0.836}&\textbf{	0.967}&	\textbf{0.992}&\textbf{	0.110}&	\textbf{0.051}&	\textbf{0.498}&	\textbf{0.174}\\
    
    \bottomrule
    
    \end{tabular}
    \caption{Comparison on NYU Depth v2 between with and without AVCL under settings of different loss functions and training data.}
    \label{tab:lossanddata}
\end{table}

\noindent\textbf{Evaluation Metrics.} For quantitative evaluation we employ several commonly used metrics: the accuracy under threshold($\delta < 1.25^i, i=1,2,3$), mean absolute relative error (rel), mean $\log 10$ error ($\log 10$), root mean squared error (rms) and rms(log).

\noindent\textbf{Implementation Details.} To fairly demonstrate the generality of the proposed AVCL, we follow the network architectures and loss functions in previous works when compared with them. For instance, when compared with DORN~\cite{dorn} we use ResNet-101~\cite{resnet} and DORN loss, while compared with Deeper Depth~\cite{deeperdepth} we use ResNet-50 and berHu loss. 
All training images are down-sampled to half size with invalid border cropped, and the predicted depth map are up-sampled to original size at the end, following ~\cite{crop1,crop2,crop3}. The hyperparameter $\mathbf{\lambda}$ is set to $(1,1,1,1,1,1)$ in all experiments. 

\vspace{0.3cm}
\hspace{-0.7cm}
\makeatletter\def\@captype{figure}\makeatother
\begin{minipage}{0.55\textwidth} 
\centering 
\includegraphics[width=\linewidth]{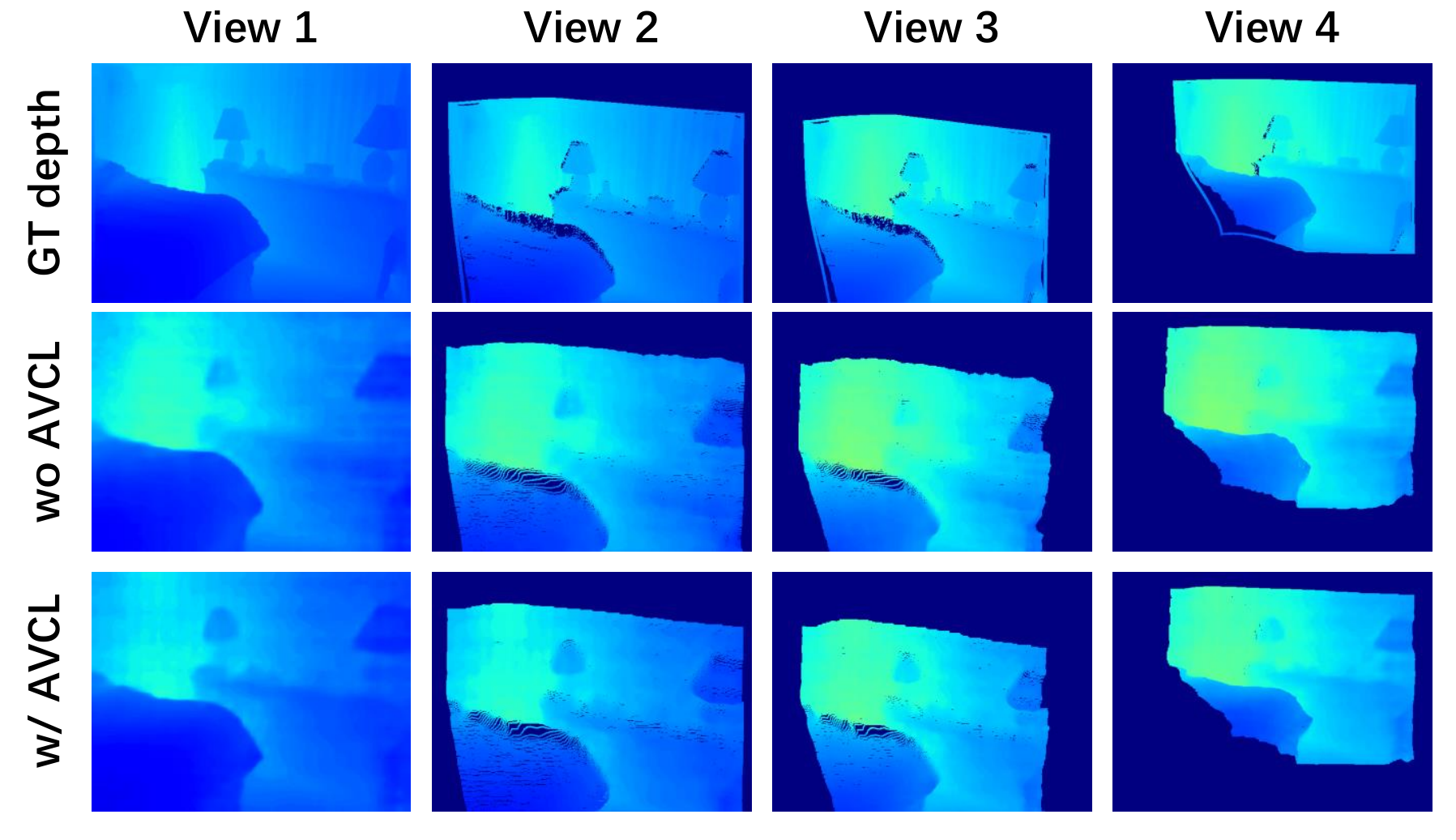}
\caption{Qualitative comparison between with and without AVCL in different views.}
\label{fig:comp}
\end{minipage} 
\makeatletter\def\@captype{table}\makeatother 
\begin{minipage}{.4\textwidth} 
\footnotesize
\centering 

\begin{tabular}{c|c|c|c|c} 

  \toprule
  \footnotesize{$\delta_1$}  &  \footnotesize{V1} & \footnotesize{V2} & \footnotesize{V3} & \footnotesize{V4} \\ 
  \hline
  \footnotesize{w/ AVCL} & \footnotesize{0.825} & \footnotesize{\textbf{0.836}} & \footnotesize{0.816} & \footnotesize{0.826} \\ 
  \footnotesize{wo AVCL} & \footnotesize{0.818} & \footnotesize{0.827} &  \footnotesize{0.808}& \footnotesize{0.820}  \\ 
  \bottomrule

\end{tabular}
\caption{Quantitative comparison between with and without AVCL in different views.}
\label{tab:consistent}
\end{minipage} 

\subsection{Experimental Results}
\label{exp:results}
In this section, we will show three significant merits of the proposed AVCL: First, with AVCL the depth generator outputs more view-consistent depth maps. Second, the demand of training data amount in AVCL is largely decreased when the same performance is reached. Last, the proposed AVCL  consistently brings performance gain when combined with state-of-the-art MDE approaches~\cite{dorn,deeperdepth}, which means AVCL is complementary to these methods. Finally, we demonstrate by an ablation study that the performance gain comes from both warping-based multi-supervision (View-Consistent Learning), and the adversarial pose generation module.

\noindent\textbf{View-Consistent.} We perform qualitative and quantitative analysis to validate that the proposed AVCL generates more view-consistent depth map. Firstly, we compare several  views of the predicted depth map with and without AVCL in Fig. \ref{fig:comp}. As it can be observed, when the network is trained with AVCL, the output depth maps have sharper object edges, and consistently presents smaller error \wrt the ground truth across different views. With fixed warping poses, we also quantitatively compute the  accuracy on the testing data, and compare in Tabel.~\ref{tab:consistent} the $\delta_1$ accuracy of these views with and without AVCL. Experimental results show that, with AVCL, the accuracy of the predicted depth map in almost all the views significantly increases, which means the predicted depth maps are more consistent across different views.

\begin{table}[]

    \centering
    \begin{tabular}{c|ccc|cccc}
    \toprule
         \multirow{2}{*}{Method} &\multicolumn{3}{c|}{Higher is better}  & \multicolumn{4}{c}{Lower is better}  \\
    \cline{2-8}
         & \small{$\delta < 1.25$}  \hspace{-0.2cm}& \hspace{-0.2cm}\small{$\delta < 1.25^2$}\hspace{-0.2cm} &\hspace{-0.2cm} \small{$\delta < 1.25^3$} \hspace{-0.2cm}&
       \small{rel}\hspace{-0.2cm} & \hspace{-0.2cm}\small{log$_10$}\hspace{-0.2cm} & \hspace{-0.2cm}\small{rms}\hspace{-0.2cm} & \hspace{-0.2cm}\small{rms$_{\log}$}\hspace{-0.2cm} \\
    \hline
    Make3D\cite{Make3D}
    &0.447	& 0.745& 0.897	& 0.349&  -   & 1.214 &-\\
    DepthTransfer\cite{depthtransfer}
    &  - 	&   -  &  - 	& 0.35 & 0.131& 0.12 &-\\
    Liu \etal\cite{discrete-continuous}
    &  - 	&   -  &  - 	& 0.335& 0.127& 1.06 &-\\
    Zhuo \etal\cite{resolution}
    &0.525	& 0.838& 0.962	& 0.305 &0.122 &1.04 &-\\
    Ladicky \etal\cite{Ladicky_2014_CVPR}
    &0.542	& 0.829& 0.941	& -  & - & - & -\\
    Wang \etal\cite{Wang_2015_CVPR}
    &0.605	& 0.890& 0.970	& 0.220  & - & 0.824 & -\\
    Eigen \etal \cite{NIPS2014_5539}
    &0.611 &0.887 &0.971 &0.215& -&0.907& 0.285\\
    Li \etal \cite{Li_2015_CVPR}
    &0.621	& 0.886& 0.968	& 0.232& 0.094& 0.821& -\\
    Liu \etal \cite{Liu}
    &0.650 &0.906 &0.976 &0.213 &0.087 &0.759& - \\
    Eigen\etal\cite{crop1}
    &0.769 &0.950 &0.988 &0.158 & -& 0.641 &0.214 \\
    Li \etal \cite{Li_2017_ICCV}
    &0.788 &0.958 &0.991 &0.143 &0.063 &0.635 &- \\
    Chakrabarti \cite{NIPS2016_6510}
    &0.806 &0.958 &0.987 & 0.149 &- &0.620 \\
    Laina \etal\cite{deeperdepth}
    &0.811 & 0.953 & 0.988 & 0.127 & 0.055 & 0.573 &0.195 \\
    MS-CRF \cite{MS-CRF}
    &0.811 & 0.954 &0.987 &0.121 &0.052 &0.586 &- \\
    DORN~\cite{dorn}
    &0.828	& 0.965& \textbf{0.992}	& 0.115 & 0.051& 0.509 & -\\
    \hline
     DORN + AVCL
    &\textbf{0.836}	& \textbf{0.967} &\textbf{ 0.992	}& \textbf{0.110}& \textbf{0.050} & \textbf{0.498} &\textbf{0.174}\\
    
    \bottomrule
    
    \end{tabular}
    \caption{Comparisons with existing monocular depth estimation methods on NYU Depth v2 dataset.}
    \label{tab:sota}
\end{table}

\noindent\textbf{Experiments with different losses.} We test with several widely-used loss functions, namely $L_1$ loss, berHu loss~\cite{deeperdepth} and DORN loss~\cite{dorn}, to demonstrate the generality of the proposed AVCL. To use berHu loss in AVCL, it is straightforward that one only needs to replace  $L_1$ loss in  $\mathcal{L}_{\texttt{dep}}$, $\mathcal{L}_{\texttt{warp}}$ and $\mathcal{L}_{\texttt{adv}}$ with berHu loss.
As for DORN loss, the case is different. DORN transforms the ordinal regression into multiple binary classification, and takes the summation of all the predicted ones as the output depth. 
Therefore, there is no such thing as a continuous depth map that can be warped and applied supervision on. To address this problem, we adopt a soft approximation of the thresholding-and-summation operation by a summation of shifted sigmoid functions $\sum_i^D sigmoid(100(x-0.5))$, where $D$ is the number of binary classifiers. In this way, the warped depth map is continuous and differentiable, and we apply $L_1$ loss on the warped branch.

Table.~\ref{tab:lossanddata} shows comparisons between different loss functions with and without AVCL. As expected, DORN loss presents better performance in terms of all the metrics than berHu loss, and the $L_1$ loss is the worst. Trained with AVCL, performances with all loss functions consistently increase, even the accuracy of the DORN baseline is already quite high. Theses results suggest that our method is complementary to these existing works on monocular depth estimation, which is reasonable because none of them considers the view-consistency.

\noindent\textbf{Experiments on small training set.} In one training pass, AVCL applys two branches of supervisions to train the depth generator, which can be understood as a kind of guided data augmentation. As a consequence, the demand of training data amount is dynamically decreased in AVCL. 
As shown in Fig.~\ref{fig:comp}, when trained with the $\frac{1}{10}$ size training set \texttt{down10}, the performance gain of AVCL is larger than trained with the entire training set. Furthermore, trained with  \texttt{down10} and AVCL, the performance is almost the same level as trained with \texttt{all} and without AVCL, or even slightly better. These results show that, with the same amount of training data, the proposed AVCL framework utilizes data more efficiently, so that the model is not prone to be overfitting.

\noindent\textbf{Comparison with existing methods.} Tabel.~\ref{tab:sota} showcases several state-of-the art monocular depth estimation methods on NYU Depth v2 dataset.
In fact, AVCL is complementary to existing methods so that the performance gain from AVCL and existing methods is addable. For instance, DORN+AVCL outperforms DORN by $0.8\%$ $\delta_1$ accuracy trained with all training data. This results may seem incremental, but one should note that when trained with $\frac{1}{10}$ data the gain is more remarkable, \ie, from $0.810$ to $0.827$ with $+1.7\%$  $\delta_1$ accuracy.

\begin{table}[]

    \centering
    \begin{tabular}{l|ccc|cccc}
    \toprule
         \multirow{2}{*}{Method} &\multicolumn{3}{c|}{Higher is better}  & \multicolumn{4}{c}{Lower is better}  \\
    \cline{2-8}
         & \small{$\delta < 1.25$}  \hspace{-0.2cm}& \hspace{-0.2cm}\small{$\delta < 1.25^2$}\hspace{-0.2cm} &\hspace{-0.2cm} \small{$\delta < 1.25^3$} \hspace{-0.2cm}&
       \small{rel}\hspace{-0.2cm}& \hspace{-0.2cm}\small{log$_{10}$}\hspace{-0.2cm} & \hspace{-0.2cm}\small{rms}\hspace{-0.2cm}& \hspace{-0.2cm}\small{rms$_{\log}$}\hspace{-0.2cm} \\
    \hline
    Baseline
    &0.817&	0.951&	0.987&	0.119& 0.062 &	0.523&	0.185\\
    Fixed pose A 
    &0.822	& 0.960& 0.989	& 0.117& 0.059&0.506& 0.177\\
    Fixed pose B
    &0.809	& 0.957& 0.984	& 0.121& 0.061 & 0.510&0.186\\
    Fixed pose C 
    &0.802	& 0.956& 0.983	& 0.124&0.062& 0.530& 0.187\\
    Random pose
    &0.825	& 0.961& \textbf{0.990}	& 0.115& 0.052 & 0.503& 0.176\\
    Adversarial pose
    &\textbf{0.836}	& \textbf{0.962}& \textbf{0.990}	&\textbf{ 0.110}&\textbf{0.050} &\textbf{0.498}& \textbf{0.174}\\
    
    \bottomrule
    
    \end{tabular}
    \caption{Comparisons with existing monocular depth estimation methods on NYU Depth v2 dataset.}
    \label{tab:abl}
\end{table}
\noindent\textbf{Ablation study.} In order to analyze the details of the proposed AVCL, we conduct several ablation studies, and results are shown in Tabel~\ref{tab:abl}. In all experiments we employ $L_1$ loss and use the \texttt{down10} training set. We start with a simple baseline without any bells and whistles (Baseline). Accordingly, we design two warping startegy to test the effectiveness of multi-view supervision: the first is to adopt a fixed warping pose (Fixed pose A-C), and the second is to randomly generate the warping pose (Random pose). Results show both fixed warping pose and  random generated pose outperforms the baseline. However, with different fixed pose, the performance gain is sensitive to the pose, ranging from $-0.015$ to $+0.005$. Randomly generated poses are superior to fixed poses, since the view point is more diverse, thus more information/supervision is introduced.
Finally, we add the adversarial pose generation module, so that the warping pose is expected to be the hardest under the recent state. Results show that the adversarial poses outperforms randomly generated poses, which proves the effectiveness and necessity of the adversarial pose generation module.

\section{Conclusion}
In this paper, we empirically find existing MDE approaches often result in sub-optimal solution that the output depth maps present inconsistent losses viewed from different directions. We therefore propose an Adversarial View-Consistent Learning framework, force the estimated depth map to be all reasonable viewed from multiple views. To this end, we design a differentiable depth map warping operation to generate multiple views of the depth map given a transformation pose, and meanwhile propose an adversarial pose generator to generate the hard poses. Experiments show learning from multiple views leads to more view-consistent predictions, and the adversarial generated poses further increases the accuracy. 

\bibliography{main}
\end{document}